\documentclass{article}
\usepackage{spconf,amsmath,graphicx,hyperref}

\usepackage{graphicx}

\usepackage{tcolorbox}
\usepackage{multirow}
\usepackage{booktabs} 
\usepackage{colortbl}
\usepackage{amsmath}
\usepackage{amssymb}

\usepackage{enumitem}

\usepackage{booktabs}
\usepackage{graphicx}

\title{PART: Progressive Alignment Representation Training
for Multilingual Speech-To-Text with LLMs}
\name{Pei Zhang$^{*13}$, Andong Chen$^{*1}$, Xi Chen$^{*12}$, Baosong Yang$^{1}$, Derek F. Wong$^{3}$, Fei Huang$^{1}$
\thanks{These authors contributed equally to this work and should
be considered co-first authors.}}
\address{$^{1}$ Tongyi Lab, Alibaba Group, $^{2}$The Chinese University of Hong Kong, \\
$^{3}$ NLP{$^2$}CT Lab, University of Macau}
\begin{document}
\ninept
\maketitle
\begin{abstract}
\vspace{0.1in}
Large language models (LLMs) have expanded from text to speech, giving rise to Speech Large Models (SLMs) that support recognition, translation, and synthesis. A key challenge is aligning speech and text representations, which becomes harder in multilingual settings. Existing methods often freeze LLM parameters and train encoders on multilingual data, but this forces cross-language convergence and limits performance. We introduce Progressive Alignment Representation Training (PART), a multi-stage and multi-task framework that separates within-language from cross-language alignment. During cross-language training, LLM parameters are dynamically activated, and text-based tasks are later introduced to enhance multilingual understanding. Experiments on CommonVoice 15, Fleurs, Wenetspeech, and CoVoST2 show that PART surpasses conventional approaches, with analysis confirming its ability to balance language-specific distinctions and cross-language generalization. These results demonstrate PART’s effectiveness and generality for multilingual speech modality alignment.
\end{abstract}
\section{Introduction}
In the research of large language models (LLMs), the scope has expanded from text to other modalities. Among them, Speech Large Models (SLMs) that use speech input and output have been widely studied and applied, showing impressive performance in tasks such as speech recognition~\cite{hu-etal-2024-vhasr}, translation~\cite{10657759,10.1109/TASLP.2024.3434425}, and speech synthesis~\cite{kim2025revival}.

The mainstream architecture of current SLMs usually consists of a pre-trained speech encoder connected to an LLM through an adapter~\cite{DBLP:journals/corr/abs-2409-17044,gaido-etal-2024-speech}. Within this framework, a central challenge is how to effectively align speech representations with the textual representations of the LLM~\cite{DBLP:conf/interspeech/HuangLGIKMP24,koneru-etal-2024-blending}. This need becomes more critical and difficult in fine-grained multilingual scenarios. 
In practice, the conventional approach often treats multilingual tasks as monolingual ones, mixing non-crosslingual tasks like automatic speech recognition (ASR) with crosslingual tasks like speech to text translation (S2TT) during training.
A common approach is to keep the LLM parameters frozen and train the speech encoder on multilingual speech data so that it aligns with the LLM input layer~\cite{du-etal-2025-making,chu2024qwen2}. However, when applied to multilingual tasks, this strategy may force audio representations of different languages to converge and place an excessive burden on the speech encoder~\cite{xu2025qwen2,tang2024salmonn}. As a result, such methods remain at the modality-level alignment, lacking deep exploration of multilingual speech-text alignment and facing performance bottlenecks in multilingual speech tasks~\cite{huang2025enhancing,nguyen2025qwen}.

We propose a \textbf{P}rogressive \textbf{A}lignment \textbf{R}epresentation \textbf{T}raining approach (\textbf{PART})
for Multilingual SLMs. In this approach, the training is divided into stages to separate within-language alignment from cross-language alignment, preventing excessive convergence of audio representations across languages. For cross-language tasks, LLM parameters are dynamically activated, allowing the speech encoder to focus on semantic mapping within each language while the LLM leverages its multilingual modeling strength. In the final stage, text-based tasks are introduced to fine-tune the LLM, further improving its ability in multilingual instruction understanding and generation. Overall, our method enables a collaborative division of labor between the speech encoder and the LLM, preserving language-specific distinctions while enhancing cross-language generalization.

Compared with traditional training methods, our multi-stage and multi-task alignment approach achieves better performance on CommonVoice 15, Fleurs, and Wenetspeech (ASR tasks), as well as CoVoST2 (S2TT task). In addition, analytical experiments show that activating the LLM in cross-language tasks effectively leverages its multilingual modeling strength, while introducing text-based tasks for fine-tuning in the final stage significantly improves multilingual ability. Overall, the results confirm the effectiveness and generality of our method for multilingual speech modality alignment. 

Our contributions are summarized as follows:
\begin{itemize}[topsep=1pt, partopsep=-1pt, itemsep=-1pt]
    \item We propose a multi-stage, multi-task alignment framework for multilingual speech-LLMs, which separates in-language alignment from multilingual alignment to better preserve language-specific features. 
    \item We design a task-dependent activation strategy that freezes the LLM in in-language ASR tasks while activating it in multilingual tasks (e.g., S2TT), allowing the audio encoder and LLM to play to their respective strengths. 
    \item We further introduce a final text-based fine-tuning stage that enhances multilingual ability, leading to improved performance across ASR and S2TT tasks. 
\end{itemize}


\section{Method}
\label{sec:method}

In this section, we first formulate the multilingual speech-to-text translation task in \ref{sec:problem}, then present the model architecture of our proposed method in \ref{sec:model}, followed by a detailed description of our progressively aligned training strategy in \ref{sec:training}.

\begin{figure*}
    \centering
    \includegraphics[width=1.0\linewidth]{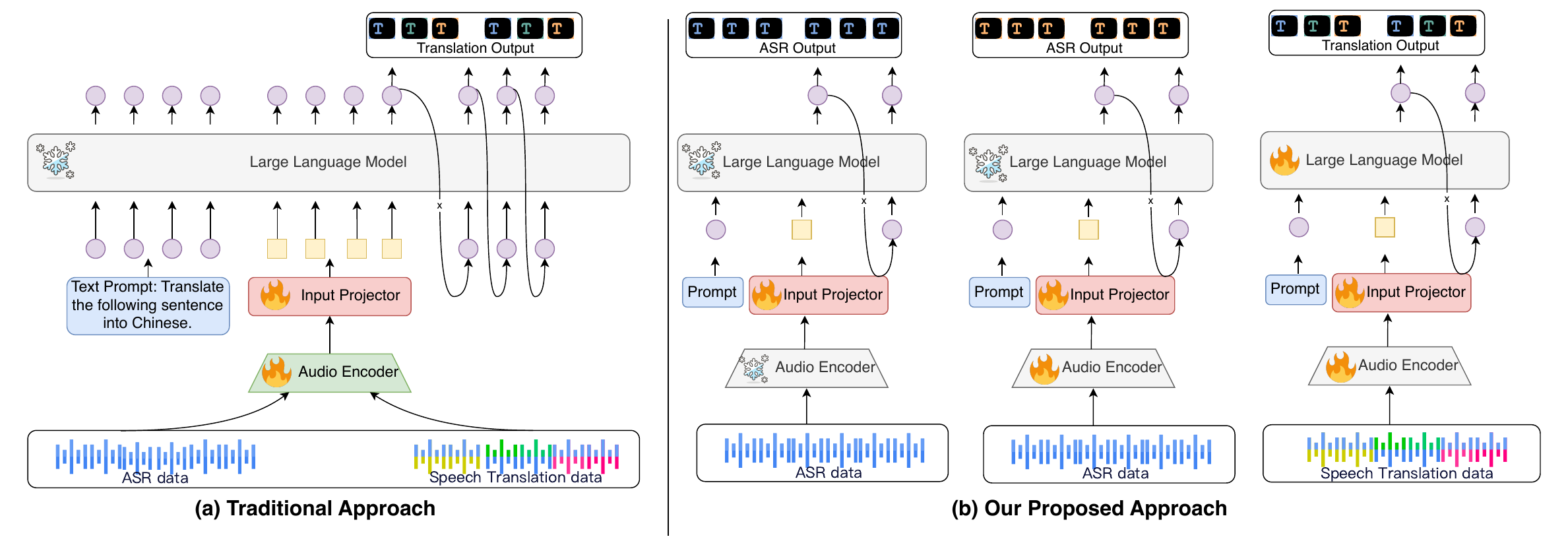}
    \caption{Our proposed training recipe \emph{v.s.} traditional training approach}
    \label{fig:recipe}
\end{figure*}

\subsection{Problem Formulation}
\label{sec:problem}
We formulate the multilingual speech-to-text task as follows: given a speech input $x$, the goal is to generate its corresponding textual output $y$. This work focuses on two distinct settings: multilingual ASR and S2TT. The language of $x$ is referred to as the speech in source language. For multilingual ASR, which is a monolingual task, the output $y$ is a transcription in the same language as $x$, trained using the monolingual dataset $\mathcal{D}_{\text{mono}}$. In contrast, speech translation is a cross-lingual task where $y$ is produced in a target language different from the source language, leveraging the cross-lingual dataset $\mathcal{D}_{\text{cross}}$.

\subsection{Model Architecture}
\label{sec:model}
As in \cite{gong2024listen,chen-etal-2024-llast}, we first extract log-mel spectrogram features $\mathbf{M}$ from the input speech signal $\mathbf{X}$ using a feature extractor. These features are then fed into a pre-trained multilingual speech encoder (parameterized by $\theta_{\text{se}}$) to obtain linguistic representations $\mathbf{X}_{\text{ling}}$. Subsequently, a lightweight adaptor module, randomly initialized with parameters $\theta_{\text{adaptor}}$ and not pretrained, projects $\mathbf{X}_{\text{ling}}$ into the embedding space of a large language model (LLM, parameterized by $\theta_{\text{llm}}$), resulting in the aligned feature representation $\mathbf{X}_{\text{align}}$. This adapted representation is then concatenated with the embedding of the instruction token $\mathbf{X}_I$, and the combined input is passed to the multilingual LLM. Finally, the LLM generates the corresponding text. Method details are shown in Figure \ref{fig:recipe}.

\subsection{Progressive Training for Multilingual SLMs}
\label{sec:training}
In contrast to existing methods, this paper contends that multilingual tasks introduce significant challenges for alignment. To address this, a progressive alignment approach is adopted, consisting of three main stages. The first two stages are responsible for gradual cross-modality alignment, while the third stage builds upon this modality-aligned foundation to fine-tune on cross-lingual tasks, thereby better leveraging the multilingual capabilities of the LLM. All three stages share the same underlying optimization objective, which is formally defined by the loss function in Eq.~\ref{eq:loss}

\begin{equation}
\begin{gathered}
\mathcal{L}_\mathcal{D} = -\mathbb{E}_{(\mathbf{X}, \mathbf{y}) \sim \mathcal{D}} \log P\bigl(\mathbf{y} \mid \mathbf{X}; \theta_{\text{se}}, \theta_{\text{adaptor}}, \theta_{\text{llm}}\bigr) \\
P(\mathbf{y} \mid \mathbf{X}) = \prod_{t=1}^{T} P\bigl(y_t \mid \mathbf{y}_{<t}, \mathbf{X}; \theta_{\text{se}}, \theta_{\text{adaptor}}, \theta_{\text{llm}}\bigr)
\end{gathered}
\label{eq:loss}
\end{equation}

\subsubsection{Stage 1: Adapter-only Within-Language Alignment}
Since both the speech encoder and the LLM are pre-trained on large-scale datasets, in the first stage, as Eq.~\ref{eq:stage1}, we fine-tune the adaptor using $\mathcal{D}_{\text{mono}}$ to perform an initial coarse-grained alignment. This provides a robust starting point for the subsequent joint optimization of multiple components.
\begin{equation}
    \theta_{\text{adaptor}}^* = \arg\min_{\theta_{\text{adaptor}}} \mathcal{L}_{\mathcal{D}_{\text{mimo}}}
\label{eq:stage1}
\end{equation}

\vspace{-0.1in}
\subsubsection{Stage 2: Within-Language Alignment with Progressive Encoder Unfreezing}
Following the first-stage fine-tuning of the adaptor, its lightweight design leads to insufficient representational capacity. To achieve more precise alignment in the second stage, as Eq.~\ref{eq:stage2}, we progressively unfreeze the speech encoder and optimize it jointly with the adaptor. More specifically, we employ a two-phase unfreezing strategy: first, the last eight layers of the speech encoder are unfrozen and fine-tuned alongside the adaptor; then, the entire speech encoder is activated for full network optimization, and this progressive activation will be discussed in Sec.~\ref{subsec:ab}. Notably, this stage continues to use only the monolingual dataset.

\begin{equation}
    \theta_{\text{adaptor}}^*, \theta_{\text{se}}^*= \arg\min_{\theta_{\text{adaptor}}, \theta_{\text{se}}} \mathcal{L}_{\mathcal{D}_{\text{mimo}}}
\label{eq:stage2}
\end{equation}

\subsubsection{Stage 3: Joint Optimization with LLM-Adaptive}
Following the alignment achieved in the first two stages, speech features and textual representations of the corresponding language are now aligned in the semantic space. We then introduce cross-lingual tasks to better leverage the multilingual capabilities of the LLM. However, due to inherent discrepancies in length between speech and text, as well as the rich diversity in speech (such as variations in speaking rate), the granularity of speech representations cannot be strictly matched to that of text. Therefore, in the third stage, as Eq.~\ref{eq:stage3}, we unfreeze the LLM and perform joint optimization of the speech encoder, adaptor, and LLM together, enhancing the model's robustness to such variations.
\begin{equation}
    \theta_{\text{adaptor}}^*, \theta_{\text{se}}^*, \theta_{\text{llm}}^*= \arg\min_{\theta_{\text{adaptor}}, \theta_{\text{se}}, \theta_{\text{llm}} }\mathcal{L}_{\mathcal{D}_{\text{mimo}} + \mathcal{D}_{\text{cross}}}
\label{eq:stage3}
\end{equation}

Through the proposed three-stage progressive alignment framework, our approach effectively bridges the gap between speech and text modalities across languages. This structured training strategy enables the model to fully leverage the LLM's inherent multilingual capabilities, facilitating robust cross-lingual transfer. As a result, the system achieves enhanced performance in multilingual speech-to-text tasks, offering improved adaptation to variability in speech while maintaining semantic coherence.

\section{Data and Training Setting}
\label{sec:exp}

In this section, we provide a detailed description of the training data sources, test datasets, and evaluation metrics in \ref{sec:data}. The model configuration and training hyperparameters are specified in \ref{sec:modelsetting}.

\subsection{Data and Settings}
\label{sec:data}

The training data includes two tasks, ASR (Automatic Speech Recognition) and S2TT (Speech-to-Text Translation). 
For ASR, there is a total of 810k hours of commercial purchases data covering 10 languages: Chinese (zh), English (en), Japanese (ja), Korean (ko), Cantonese (yue), German (de), French (fr), Russian (ru), Spanish (es), and Italian (it). 
The S2TT task data encompasses 434k hours, covering language pairs such as zh-en, ja-en, de-en, fr-en, es-en, it-en, ru-en, as well as en-zh, en-ja, en-de, en-sv, en-id, and en-ar. 
The main sources of S2TT data are: 1) open-source datasets like CoVoST~\cite{wang-etal-2020-covost}, TED-LIUM~\cite{rousseau-etal-2012-ted}, and MuST-C~\cite{di-gangi-etal-2019-must}; 2) constructing S2TT data by translating ASR transcripts into target languages and The rest comes from commercial purchases of 48k hours.

For model capability evaluation, the ASR task includes four types of test sets: Librispeech (human-read audiobooks)~\cite{7178964}, CommonVoice 15 (crowdsourced speech)~\cite{ardila-etal-2020-common}, Fleurs (human-read Wikipedia)~\cite{conneau2023fleurs}, and Wenetspeech (audio from YouTube and podcasts)~\cite{zhang2022wenetspeech}. CER (Character Error Rate)~\cite{james2024advocating} is used for zh, ja, ko, yue, while WER (Word Error Rate)~\cite{jelinek1998statistical} is used for other languages, all combined with Whisper~\cite{radford2023robust} normalizer post-processing. 
The evaluation of translation tasks uses primarily CoVoST2~\cite{wang2020covost}, which is based on Common Voice, using BLEU scores for the evaluation, with character-based tokenization for Chinese and Japanese, and the 13a tokenizer for other languages. 
Details are elaborated in Table~\ref{tab:benchmark}.

\begin{table}[!ht]
\centering
\caption{Benchmark datasets for ASR and S2TT. “xx→en” means non-English source to English target; “en→xx” means English source to non-English target.}

\resizebox{\columnwidth}{!}{%
\begin{tabular}{lllp{3.2cm}l}
\toprule
\textbf{Task} & \textbf{Test Data} & \textbf{Domain} & \textbf{Languages} & \textbf{Metric} \\
\midrule
\multirow{4}{*}{ASR} 
 & LibriSpeech (test-clean/other) & read & en & WER \\
 & Wenetspeech (test-net/meeting) & YouTube/Podcast & \textit{zh} & CER \\
 & Fleurs & Wikipedia (read) & \textit{zh, en, ja, ko, yue,}\\ 
 &  &  & \textit{de, fr, ru, es, it} & CER/WER \\
 & Common Voice 15 & crowdsourcing & \textit{zh, en, ja, ko, yue,}\\
 &  &  & \textit{de, fr, ru, es, it} & CER/WER \\
\midrule
S2TT & CoVoST2 & crowdsourcing & \textit{xx→en: zh, ja, de, fr, es, it, ru}\\
 &  &  & \textit{en→xx: zh, ja, de, sv, id, ar} & BLEU \\
\bottomrule
\end{tabular}%
}
\label{tab:benchmark}
\end{table}

\vspace{-0.1in}
\subsection{Model Specifications}
\label{sec:modelsetting}
The speech audio encoder is initialized with the SenseVoice-large encoder~\cite{an2024funaudiollm}, approximately 700M in size. 
The LLM is based on Qwen2.5~\cite{Yang2024Qwen25TR} enhanced with multilingual continuing pre-training in advance. 
The Adapter consists of two transformer layers and one CNN layer. 
We experiment with two sizes of the Qwen2.5 model, 1.5B and 7B parameters, corresponding to PART-2B and PART-8B. Training is performed using 256 NVIDIA A800 GPUs for three epochs, and inference is performed using greedy decoding.



\section{Experiment Result}
In this section, we conduct a comprehensive set of experiments to evaluate the proposed method. In Sec.~\ref{subsec:main_res}, we compare our approach with several state-of-the-art methods. Sec.~\ref{subsec:pt} provides a detailed analysis of the contribution of our two novel designs. Finally, in Sec.~\ref{subsec:ab}, we perform ablation studies to better understand the role of progressively unfreezing strategy in stage 2.
\subsection{Main Result}
\label{subsec:main_res}

\textbf{For Multilingual Automatic Speech Recognition:} The experimental results are shown in Table \ref{asr_result}. On LibriSpeech, Wenetspeech, Fleurs, and Common Voice 15, PART consistently outperforms the two-stage baseline and remains competitive with larger SLMs. For example, on Fleurs it reduces the average WER from 6.35 to 3.73 ($\downarrow$ 41\%), and on Common Voice 15 from 9.18 to 6.29 ($\downarrow$ 32\%). These results show that the progressive alignment strategy preserves language-specific features while the task-dependent activation mechanism strengthens cross-language robustness, leading to clear advantages in multilingual ASR.

\begin{table*}[!ht]
\centering
\caption{ASR main results (WER, \%). Bold numbers indicate the best results in each column. “–” means the model does not support the corresponding language or dataset.}

\resizebox{\textwidth}{!}{
\begin{tabular}{l c cc cc cccccccccc cccccccccc}
\toprule
\textbf{Models} & \textbf{Params} 
& \multicolumn{2}{c}{\textbf{Librispeech}} 
& \multicolumn{2}{c}{\textbf{Wenetspeech}} 
& \multicolumn{10}{c}{\textbf{Fleurs}} 
& \multicolumn{10}{c}{\textbf{Common Voice 15}} \\ 
\cmidrule(lr){3-4} \cmidrule(lr){5-6} \cmidrule(lr){7-16} \cmidrule(lr){17-26}
& & clean & other & meeting & net 
& zh & en & ja & ko & yue & de & fr & ru & es & it 
& zh & en & ja & ko & yue & de & fr & ru & es & it \\ 
\midrule
SALMON & 14B & 2.1 & 4.9 & – & – & – & – & – & – & – & – & – & – & – & – & – & – & – & – & – & – & – & – & – & – \\
MinMo & 8B & 1.7 & 3.9 & 6.8 & 7.4 & \textbf{3.0} & \textbf{3.8} & 3.8 & 2.9 & 4.3 & 5.2 & 5.5 & 6.2 & 3.4 & 3.5 & 6.3 & 7.9 & 13.4 & 6.6 & 6.4 & 6.6 & 8.5 & 7.0 & 5.0 & 6.1 \\
Qwen2-Audio & 8B & \textbf{1.6} & 3.6 & 8.1 & 9.5 & 7.5 & 5.1 & 10.4 & 10.6 & 4.1 & 10.5 & 9.4 & 23.2 & 7.3 & 6.7 & 6.9 & 8.6 & 13.5 & 17.5 & 5.9 & 7.6 & 9.6 & 16.8 & 5.7 & 6.8 \\
Qwen2.5-Omni & 8B & 1.8 & \textbf{3.4} & \textbf{5.9} & 7.7 & \textbf{3.0} & 4.1 & – & – & – & – & – & – & – & – & \textbf{5.2} & \textbf{7.6} & – & – & – & – & \textbf{7.5} & – & – & – \\
\midrule
PART & 2B & 2.0 & 4.2 & 7.9 & 7.2 & 4.2 & 5.0 & 3.6 & 3.1 & 4.3 & 5.6 & 6.5 & 6.9 & 3.8 & 4.2 & 7.0 & 9.6 & 10.2 & 5.6 & 5.8 & 6.4 & 9.3 & 7.6 & 6.8 & 5.8 \\
PART & 8B & 1.7 & 3.8 & 7.5 & \textbf{6.8} & 3.9 & 4.0 & \textbf{3.1} & \textbf{2.5} & \textbf{3.7} & \textbf{4.4} & \textbf{4.7} & \textbf{5.2} & \textbf{2.8} & \textbf{3.0} & 6.4 & 8.5 & \textbf{9.7} & \textbf{4.9} & \textbf{5.5} & \textbf{5.1} & \textbf{7.5} & \textbf{5.6} & \textbf{5.1} & \textbf{4.6} \\
\bottomrule
\end{tabular}
}
\label{asr_result}
\end{table*}

\textbf{For Multilingual Speech-to-Text Translation:} The experimental results are shown in Table \ref{mt_result}. On CoVoST2 across both xx2en and en2xx directions, PART also demonstrates significant advantages. PART-8B achieves the highest BLEU scores on most language pairs, with particularly large gains in low-resource directions such as en→sv, en→id, and en→ar, where it surpasses the two-stage baseline by 3–8 BLEU. This improvement stems from the third-stage introduction of text-based tasks, which, building on modality alignment, further unlocks the LLM’s cross-lingual generation ability, enabling the model to better capture semantics and produce fluent translations in complex multilingual scenarios. At the same time, compared with Whisper-large-v2 and MinMo, PART maintains stable advantages in high-resource languages such as German and French, showing that the method not only addresses cross-lingual alignment challenges but also achieves general improvements in multilingual generation.

\begin{table*}[!ht]
\centering
\caption{S2TT main results (BLEU). Bold numbers indicate the best results in each column. “–” means the model does not support the corresponding language.}

\resizebox{\textwidth}{!}{
\begin{tabular}{l c ccccccc cccccc}
\toprule
\textbf{Model} & \textbf{Params} 
& \multicolumn{7}{c}{\textbf{xx2en}} 
& \multicolumn{6}{c}{\textbf{en2xx}} \\
\cmidrule(lr){3-9} \cmidrule(lr){10-15}
& & zh & ja & de & fr & es & it & ru 
  & zh & ja & de & sv & id & ar \\
\midrule
Whisper-large-v2 & 1.6B & 18.0 & 26.1 & 36.3 & 36.4 & 40.1 & 30.9 & – 
& – & – & – & – & – & – \\
Speech-LLaMA & 7B & 12.3 & 19.9 & 27.1 & 25.2 & 27.9 & 25.9 & 36.8 
& – & – & – & – & – & – \\
SALMON & 14B & – & – & – & – & – & – & – 
& 33.1 & 22.7 & 18.6 & – & – & – \\
MinMo & 8B & 26.0 & 28.9 & 39.9 & 41.3 & \textbf{43.3} & \textbf{40.6} & 48.6 
& 46.7 & 35.1 & – & – & – & – \\
Qwen2-Audio & 8B & 24.4 & 20.7 & 35.2 & 38.5 & 40.0 & 36.3 & – 
& 45.2 & 28.8 & 29.9 & – & – & – \\
Qwen2.5-Omni & 8B & \textbf{29.4} & – & 37.7 & – & – & – & – 
& 41.4 & – & 30.2 & – & – & – \\
LLaST & 2B & 19.2 & 24.2 & 36.8 & 41.2 & 43.2 & 39.3 & – 
& – & – & – & – & – & – \\
\midrule
PART & 2B & 23.2 & 26.8 & 37.9 & 39.2 & 40.9 & 37.3 & 46.5 
& 42.4 & 43.8 & 30.8 & 15.7 & 32.3 & 16.6 \\
PART & 8B & 27.0 & \textbf{30.0} & \textbf{40.8} & \textbf{42.3} & 42.7 & 39.7 & \textbf{50.9} 
& \textbf{46.8} & \textbf{47.2} & \textbf{35.0} & \textbf{25.8} & \textbf{37.5} & \textbf{22.4} \\
\bottomrule
\end{tabular}
}
\label{mt_result}
\end{table*}

\vspace{-0.1in}
\subsection{Analysis of Each Stage in Progressive Training}
\label{subsec:pt}

\begin{figure}[!ht]
  \centering
  \includegraphics[width=\linewidth]{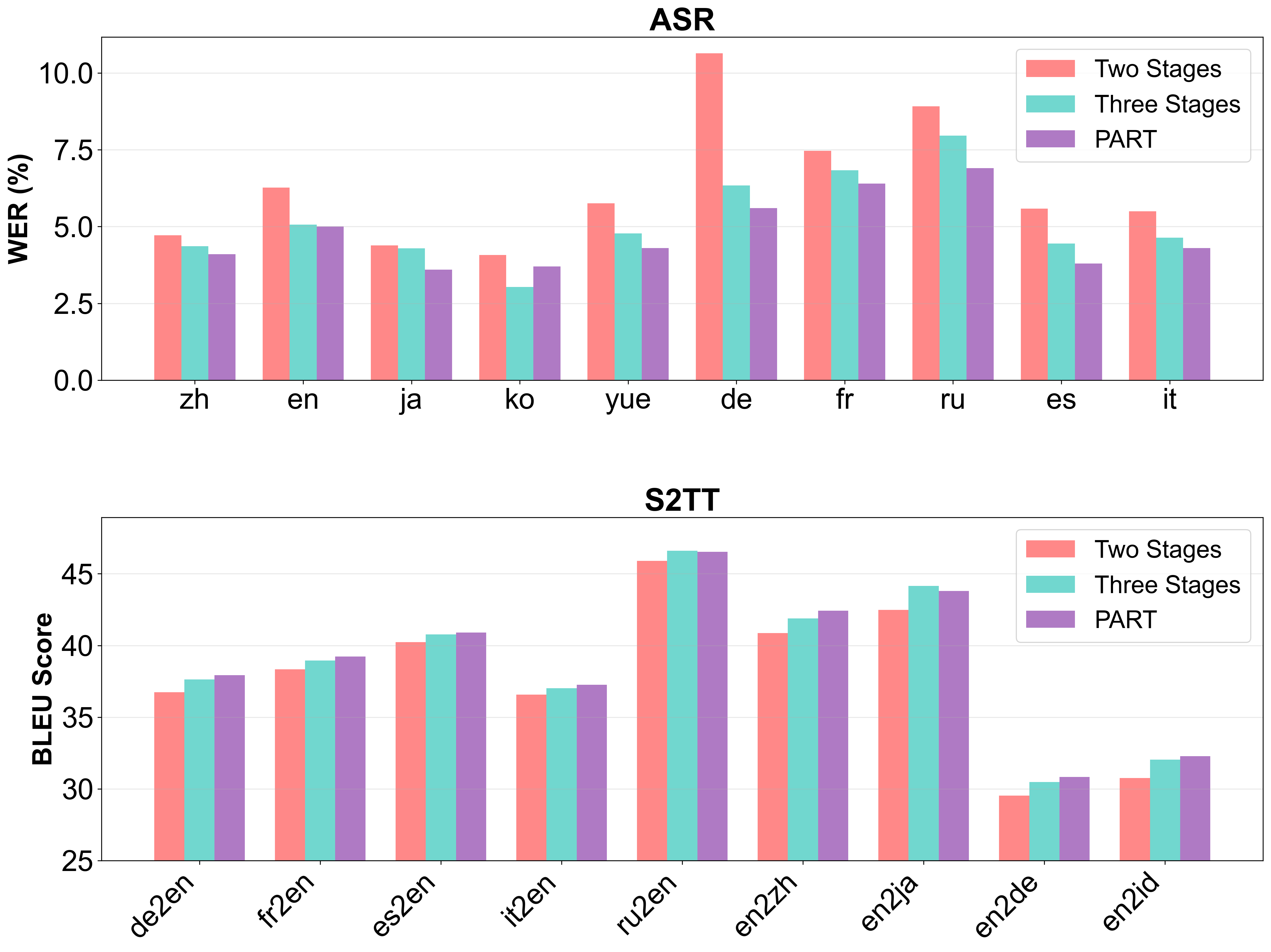}
  \caption{Comparing WER and BLEU at the progressive training stage}
  \label{fig:analysis_each_stage}
  \vspace{-0.1in}
\end{figure}

To evaluate the effectiveness of our progressive training strategy, we conducted experiments using mixed ASR and S2TT data under two configurations: two-stage training and three-stage training. The optimization settings and methodologies for both setups were consistent with those described in the Method section. We evaluated ASR performance using Word Error Rate (WER) on the FLEURS test set and S2TT performance using BLEU score on the CoVoST2 test set. As shown in Fig.~\ref{fig:analysis_each_stage}, the model trained with three stages significantly outperformed the two-stage model on both tasks, achieving lower WER and higher BLEU scores. These results underscore the superiority of our progressively fine-grained training strategy.

In addition, we verified that our proposed approach of aligning the model first and then fine-tuning it on cross-lingual tasks contributes to better convergence. As shown in Fig.~\ref{fig:analysis_each_stage}, the three-stage training method uses both ASR and S2TT tasks across all three stages, while PART (Progressive Alignment then Tuning) trains the first two stages only on ASR data and the final stage on both ASR and S2TT tasks. PART achieves lower WER and higher BLEU scores on nearly all subsets. These results confirm that performing alignment before cross-lingual fine-tuning offers a more effective training paradigm for multilingual tasks, as it reduces the language model's confusion over language-specific speech characteristics.
\vspace{-0.1in}
\subsection{Ablation experiments}
\label{subsec:ab}

To isolate and evaluate the impact of progressively unfreezing the speech encoder on both ASR and S2TT tasks, we conducted a controlled ablation study. Both the baseline ("full") and our progressive method ("last8-full") were identically trained on the combined dataset $\mathcal{D}{\text{mimo}} + \mathcal{D}{\text{cross}}$ following a two-stage protocol: first fine-tuning the adapter modules, followed by joint fine-tuning of the speech encoder and adapter. The sole difference lies in the second stage; the "full" baseline unfreezes and fine-tunes the entire speech encoder at once, while our "last8-full" strategy progressively unfreezes it (last 8 layers first, then the full encoder). As evidenced in Table~\ref{tab:ab}, our progressive approach yields a consistent improvement, reducing average WER on FLEURS by 0.1 and increasing  average BLEU on CoVoST2 by 0.3. These gains confirm that a gradual unfreezing strategy, within this framework, more effectively facilitates alignment learning compared to full-encoder fine-tuning.

\vspace{-0.1in}
\begin{table}[!ht]
\centering
\caption{Ablation study on Stage-2 fine-tuning strategies, comparing full encoder unfreezing against progressive unfreezing (last8→full). Performance is measured by average WER on 10 languages of FLEURS and average BLEU score on 13 language pairs of CoVoST2.}
\vspace{0.05in}
\resizebox{0.7\columnwidth}{!}{%
\begin{tabular}{lll}
\toprule
\textbf{Finetune Paradigm in Stage 2} & \textbf{WER(\%)} & \textbf{BLEU}\\
\midrule
last8-\>full & 6.4 & 31.7 \\
\midrule
full & 6.3 & 32.0 \\
\bottomrule
\end{tabular}%
}

\label{tab:ab}
\end{table}

\vspace{-0.1in}

\section{Conclusion}
This work proposes PART, a staged, task-dependent training paradigm that decouples intra-language alignment from cross-lingual alignment. The staged approach preserves language specificity while fully exploiting the LLM’s multilingual modeling capacity, avoiding over-convergence of speech representations in multilingual settings. Large-scale evaluations show that PART achieves significant gains on ASR and S2TT. Mechanistic analyses and ablations validate the method’s effectiveness and its marked improvements in robustness to Language Identification (LID) prompts. Overall, PART provides a general and transferable paradigm for multilingual speech–text alignment, and offers valuable insights for future optimization of multilingual speech-language models.

\bibliographystyle{IEEEbib}
\bibliography{strings,refs}





\end{document}